\PassOptionsToPackage{table}{xcolor}
\documentclass[runningheads]{llncs}
\usepackage[none]{hyphenat}
\usepackage{graphicx}
\usepackage{subfiles}
\usepackage{array,booktabs}
\usepackage[final]{pdfpages}
\usepackage{multirow}
\usepackage{array}
\usepackage[misc]{ifsym}

\usepackage{xcolor}

\usepackage[linesnumbered,vlined,ruled]{algorithm2e}
\usepackage{algpseudocode}

\usepackage{tikz}

\usetikzlibrary{arrows}
\usetikzlibrary{positioning,decorations.pathreplacing,shapes}
\usepackage[english]{babel}
\usepackage{microtype}
\usepackage{amsmath}
\usepackage{amssymb}
\tikzstyle{arrow} = [thick,->,>=stealth, shorten <=3pt, shorten >=3pt]

\begin{document}
\title{A Generic and Model-Agnostic Exemplar Synthetization Framework for Explainable AI} 
\author{Antonio Barbalau$^{1,3}$ {\Letter} \and Adrian Cosma$^{2, 3}$ \and Radu Tudor Ionescu$^1$ \and Marius Popescu$^{1,3}$}
\authorrunning{A. Barbalau et al.}
\institute{University of Bucharest, {\Letter} : \email{abarbalau@fmi.unibuc.ro}, \and University Politehnica of Bucharest, $^3\;$ Sparktech Software}



\titlerunning{A Generic and Model-Agnostic Exemplar Synthetization Framework}
\tocauthor{A. Barbalau, A. Cosma, R.T. Ionescu, M. Popescu}
\toctitle{A Generic and Model-Agnostic Exemplar Synthetization Framework for Explainable AI}
 
\maketitle

\begin{abstract}
With the growing complexity of deep learning methods adopted in practical applications, there is an increasing and stringent need to explain and interpret the decisions of such methods. In this work, we focus on explainable AI and propose a novel generic and model-agnostic framework for synthesizing input exemplars that maximize a desired response from a machine learning model. To this end, we use a generative model, which acts as a prior for generating data, and traverse its latent space using a novel evolutionary strategy with momentum updates. Our framework is generic because $(i)$ it can employ any underlying generator, e.g. Variational Auto-Encoders (VAEs) or Generative Adversarial Networks (GANs), and $(ii)$ it can be applied to any input data, e.g. images, text samples or tabular data. Since we use a zero-order optimization method, our framework is model-agnostic, in the sense that the machine learning model that we aim to explain is a black-box. We stress out that our novel framework \emph{does not} require access or knowledge of the internal structure or the training data of the black-box model. We conduct experiments with two generative models, VAEs and GANs, and synthesize exemplars for various data formats, image, text and tabular, demonstrating that our framework is generic. We also employ our prototype synthetization framework on various black-box models, for which we only know the input and the output formats, showing that it is model-agnostic. Moreover, we compare our framework (available at \url{https://github.com/antoniobarbalau/exemplar}) with a model-dependent approach based on gradient descent, proving that our framework obtains equally-good exemplars in a shorter time.
\end{abstract}

\section{Introduction}
Due to the rise of deep learning~\cite{LeCun-Nature-2015} in recent years, scientists and engineers have developed solutions based on deep learning to solve almost every machine learning task in production-ready systems. While deep learning models obtain impressive accuracy levels~\cite{Devlin-NAACL-2019,He-CVPR-2016,Krizhevsky-NIPS-2012,Ren-NIPS-2015}, surpassing even human-level performance for many tasks~\cite{Cozma-ACL-2018,Georgescu-Access-2019}, their inherent complexity transforms them into opaque decision systems.
Critical processes that deal with potentially sensitive information in areas such as finance, medicine, security and justice have become essentially black boxes, with the underlying logic being too complex even for data scientists and inaccessible to the end-users. Deep learning models require training data usually generated and annotated by humans, thus containing various biases, including discriminatory views on race and gender. Hence, models trained on biased data will inherit these biases and, in turn, will make decisions that are unfair or socially unacceptable~\cite{Caliskan-Science-2017}. Another potential problem of such highly complex decision systems is the chance of inadvertently making correct decisions, but for the wrong reasons. A popular example here is an image of a wolf being classified correctly, but only because of the snowy background \cite{Ribeiro-KDD-2016}. While this is a harmless example, the same kind of decisions resulting from spurious correlations from large amounts of data could potentially have a great negative impact on human lives.

In this context, explainable AI, a field that studies how artificial intelligence (AI) methods and techniques can be understood by human experts, gained a lot of attention recently. While there are many types of explanations that an explanatory method could provide \cite{Adadi-Access-2018,Guidotti-CS-2018}, including rule extraction and outcome prediction, we chose to focus on explanation by exemplar generation. Prototypical examples (exemplars), which can describe a complex underlying data distribution, can offer meaningful insights about the behavior of a model, when a simple explanation is hard to extract. Prototype selection methods \cite{Bien-AAS-2011,Chen-NeurIPS-2019,Gurumoorthy-ICDM-2019,Mahendran-CVPR-2015,Yeh-NIPS-2018} return examples that are representative for a set of similar instances. An exemplar can be one of the instances observed in the training data set \cite{Bien-AAS-2011,Gurumoorthy-ICDM-2019,Yeh-NIPS-2018}, or it can be an artificially-generated example in the data space \cite{Chen-NeurIPS-2019,Mahendran-CVPR-2015}.

Current explainable AI approaches mainly consider the glass-box scenario of highly-complex  deep  learning  models,  with  no  restriction  towards accessing the models’ weights. For example, Nguyen et al. \cite{Nguyen-NIPS-2016} applied gradient descent to back-propagate through the model in order to synthesize the preferred inputs for neurons, which would not be possible without knowing the weights. We consider the more realistic case in which we have no information about the weights or other internal components of the model. Our framework is only allowed to inspect the input format and the output predictions. 
This strict definition of black-box models enables us to explain just about any machine learning model, not only deep learning models trained with gradient descent. In other words, our framework is \emph{model-agnostic}.

While related methods obtain exemplars for image classification models~\cite{Guidotti-ECML-2019,Nguyen-NIPS-2016}, we propose a more generic framework that it is not tied to a particular data modality -- be it image, text, structured or tabular. Our framework is also agnostic with respect to the underlying generative model used to synthesize exemplars -- be it a Variational Auto-Encoders (VAE) \cite{Kingma-ICLR-2014} or a Generative Adversarial Network (GAN) \cite{Goodfellow-NIPS-2014}. To this end, we consider our framework as \emph{generic}.

\begin{figure}[!t]
    \centering
    \includegraphics[width=0.99\linewidth]{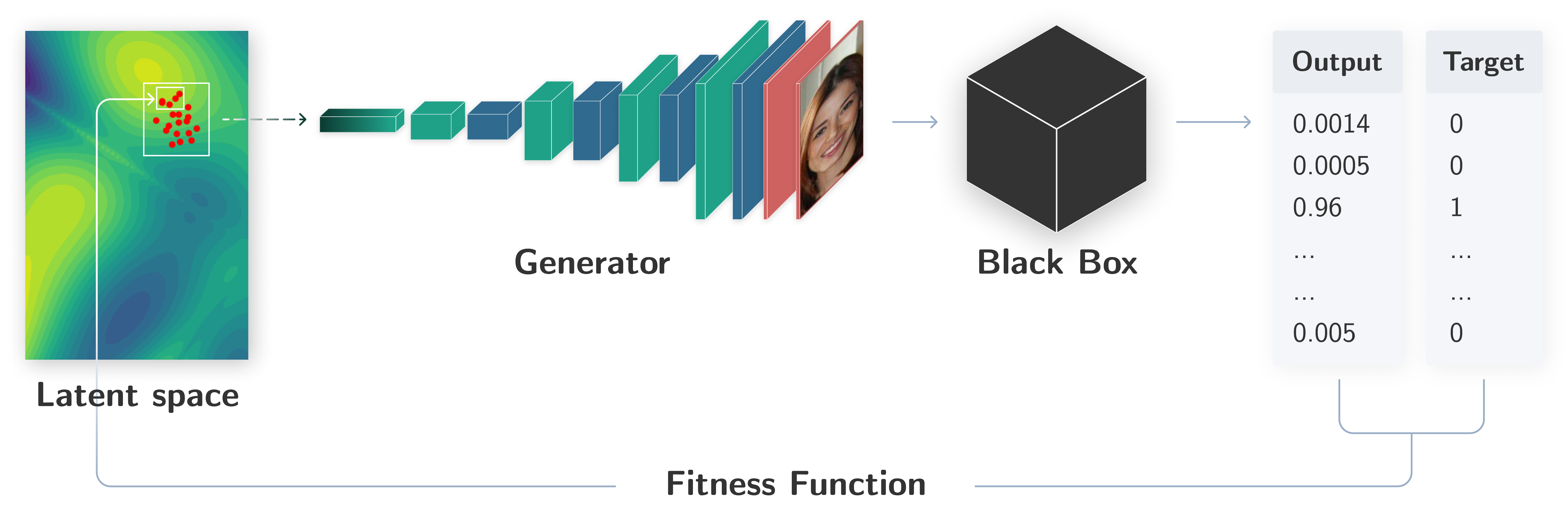}
    \caption{General diagram of our exemplar generation method. Given a black-box classifier trained on some data set, we first train a generative model (VAE / GAN) on a disjoint data set, not necessarily containing the same class distribution. 
    We then traverse the latent space of the generator, employing an evolutionary strategy, such that the black-box model provides a certain prediction (chosen beforehand) when the generated exemplar is given as input to the black-box model. Best viewed in color. Patent Pending No. 63053589.}
    \label{fig_framework_diagram}
\end{figure}

To our best knowledge, we are the first to propose a generic and model-agnostic explainable AI framework to synthesize exemplars that exhibit a high response with respect to the output of a black-box model. We exploit the structured latent space of the underlying generative model to progressively search for latent codes that can accurately explain a particular class or combination of classes, as learned by the model. We employ a novel evolutionary strategy with momentum updates as our search policy, as it was proven that evolutionary algorithms \cite{Salimans-ArXiv-2017} represent an efficient way for black-box optimization, relieving us from the need to propagate gradients. Our framework is illustrated in Figure \ref{fig_framework_diagram}.

We conduct experiments with two generative models, VAEs and GANs, to synthesize exemplars for various data formats, namely image, text and tabular, demonstrating that our framework is generic. We also employ our prototype synthetization framework on various black-box models, e.g. Random Forest or neural networks, for which we only know the input and the output formats, showing that it is model-agnostic. We present experiments showing that our framework can also generalize to classes unseen by the generator. Moreover, we compare our framework with a model-dependent approach \cite{Nguyen-NIPS-2016} based on gradient descent optimization, demonstrating that our framework converges to equally-good exemplars in a shorter amount of time.


\section{Related Work}
\label{sec_related}
Explainable AI methods \cite{Adadi-Access-2018,Guidotti-CS-2018} for black-box models have gained significant attention in recent years, as bias in data and model training \cite{Bolukbasi-NIPS-2016,Caliskan-Science-2017} have resulted in regulatory actions from the European Union \cite{Goodman-AIM-2017}, restricting the use of decision-making machine learning models without an explanatory component. Therefore, explainable AI is an area of research of utter importance, with open issues ranging from thorough testing and regulatory compliance \cite{Mittelstadt-FAccT-2019} to finding what kind of explanations are best suited to answer questions of fairness.

Explainable AI methods can be classified into different taxonomies based on various criteria, e.g. as global or local methods, as model-specific or model-agnostic methods and so on \cite{Adadi-Access-2018,Guidotti-CS-2018}. We hereby focus on methods that are closer to our own, i.e. on model-agnostic or exemplar-based methods.

We note that deep neural networks are commonly regarded as complex models, their decisions being hard to understand due to the hierarchical non-linear structure. Local explainable AI methods \cite{Karpathy-ICLR-2016,Li-NAACL-2016,Lundberg-NIPS-2017,Mullenbach-NAACL-2018,Ribeiro-KDD-2016,Selvaraju-ICCV-2017,Wiegreffe-EMNLP-2019,Zhou-CVPR-2016} deal with explaining a particular decision, i.e. the decision provided for a certain input example. Some of these methods \cite{Lundberg-NIPS-2017,Ribeiro-KDD-2016} employ a directly-interpretable surrogate model that is trained on the vicinity of an input sample, by modifying or occluding input features. 
Access to the network's weights enables the back-propagation of gradients, leading to saliency-based methods such as CAM \cite{Zhou-CVPR-2016}, Grad-CAM \cite{Selvaraju-ICCV-2017} and Grad-CAM++ \cite{Chattopadhay-WACV-2018}. Unlike these local explainable AI methods, we propose a framework that does not look inside the models, i.e. the models are complete black boxes. Furthermore, our approach provides generic (non-local) explanations by synthesizing exemplars not tied to a certain data sample.

Exemplar-based methods \cite{Bien-AAS-2011,Chen-NeurIPS-2019,Gurumoorthy-ICDM-2019,Mahendran-CVPR-2015,Yeh-NIPS-2018}, which offer a convenient way to communicate meaningful insights about the behavior of a model in situations where a direct explanation is hard to extract, are more closely related to our framework. Some exemplar-based methods select prototypical examples from the training data set \cite{Bien-AAS-2011,Gurumoorthy-ICDM-2019,Yeh-NIPS-2018}. For example, Gurumoorthy et al. \cite{Gurumoorthy-ICDM-2019} proposed a method aimed at describing the data distribution through case-based reasoning, while Chen et al. \cite{Chen-NeurIPS-2019} presented an approach that selects relevant samples from the data set that have contributed to a decision of the model. Different from methods selecting exemplars from the data set, we introduce a framework that generates realistic examples without having access to the data set used to train the black box.

Synthesizing artificial (unrealistic) examples that provide maximal responses for a particular network component can shed light on the preferences and biases of a trained model. Indeed, methods \cite{Mahendran-IJCV-2016,Simonyan-ICLR-2014} of visualizing a convolutional neural network represent a popular way of understanding its behavior. However, such methods are based on different variations of gradient ascent, requiring access to the internal weights of the model. Unlike such methods \cite{Mahendran-IJCV-2016,Simonyan-ICLR-2014}, we can produce realistic examples while treating the model as a black box.

We identified two works~\cite{Guidotti-ECML-2019,Nguyen-NIPS-2016} that are very closely related to our approach. Our method is similar to that of Nguyen et al. \cite{Nguyen-NIPS-2016} because our method, as much as theirs, requires a deep generative model to synthesize realistic images. Without a deep generative model to act as a realistic image prior, there is a high chance that preferred inputs could end up being unrealistic. In \cite{Nguyen-CVPR-2015}, the authors already proved that neural networks output high responses to texture-like images that were generated using genetic algorithms, with little to no resemblance to natural images. Nguyen et al. \cite{Nguyen-NIPS-2016} applied gradient descent to back-propagate through the model in order to synthesize preferred inputs for neurons. Being base on gradient descent, their method requires access to the model's weights. Different from Nguyen et al. \cite{Nguyen-NIPS-2016}, we consider the more realistic case in which we have no information about the weights or other internal components of the model. Our framework is only allowed to inspect the input format and the output predictions. 
This strict definition of black-box models enables us to explain just about any machine learning model, not only deep learning models trained with gradient descent. Without access to the gradients, our method generates exemplars through a  novel evolutionary strategy with momentum updates.

Focusing on image classification, Guidotti et al. \cite{Guidotti-ECML-2019} presented an approach to explain the decisions of black-box models for a given input sample. Different from Guidotti et al. \cite{Guidotti-ECML-2019}, we show that our method is applicable to different data types, namely to images, text samples and tabular data. We also show that our method works with various generators, namely VAEs and GANs. While Guidotti et al. \cite{Guidotti-ECML-2019} focus on explaining single instances, we focus on explaining output class probabilities, i.e. our exemplars are not tied to input data samples. All in all, we consider that there are significant differences between our method and that of Guidotti et al. \cite{Guidotti-ECML-2019}.


\section{Method}
\label{sec_method}
Given a black-box classification model $C:D \rightarrow \mathbb{R}^n$ and a generative model $G: \mathbb{R}^l \rightarrow D$ able to sample from a data set included in $D$, we aim to traverse the latent space of $G$ using a gradient-free optimization method, namely an evolutionary strategy with momentum updates, in order to synthesize exemplars for which $C$ provides a certain desired output $y \in \mathbb{R}^n$. Here, $n$ is the number of classes, $l$ is the size of the embedding space of the generator, and $D$ is the data space, which depends on the input data type, e.g. for images $D=[0,255]^{h\times w}$, where $h$ and $w$ are the height and the width of an input image, respectively. In our framework, we impose no restrictions upon the prediction model $C$. 
We require no access or knowledge of the internal structure of $C$, i.e. $C$ is a black box. As generator, we can use any model that takes as input a noise vector $z$ and outputs a corresponding data sample, including Variational Auto-Encoders, Generative Adversarial Networks and Auto-Regressive models. Given the target prediction $y \in \mathbb{R}^n$, we optimize an encoding $z \in \mathbb{R}^l$ such that $\hat{y} = C(G(z))$ is optimally close to $y$, i.e. $\hat{y} \approx y$. The objective for our optimization problem can be formally expressed as follows:
\begin{equation}\label{eq_objective}
V(y,C,G) = \min_{z} \sum_{i=1}^{n}(C(G(z)) - y_i)^2 = \min_{z} \sum_{i=1}^{n}(\hat{y}_i - y_i)^2.
\end{equation}

Nguyen et al. \cite{Nguyen-NIPS-2016} employed gradient descent to optimize the objective defined in Eq.~\eqref{eq_objective}, by back-propagating though the classification model $C$. However, we assume that access to the internal structure or the weights of the model is not granted, i.e. the model $C$ is a black box. Furthermore, we do not impose any architectural restrictions over the model, i.e. $C$ needs not be a neural network. Even if access to the analytical gradients is not provided due to the black-box nature of the model $C$, one can still compute the numerical gradients and search for a $z$ that minimizes the objective defined in Eq.~\eqref{eq_objective}, using gradient descent. However, since $z$ belongs to an $l$-dimensional space, computing the numerical gradients for each component in $z$ requires $l$ forward passes through the model $C$, which is inefficient in comparison to our evolutionary approach. We show in our experiments that we can synthesize exemplars with a confidence greater than $95\%$ with less than $l$ model calls (forward passes) on average. Additionally, we show that our evolutionary approach provides better exemplars and converges faster than gradient descent, even when analytical gradients are available for the classification model, i.e. $C$ becomes a glass-box model as in \cite{Nguyen-NIPS-2016}.

\SetKwProg{MyStruct}{struct}{ contains}{end}
\SetKwRepeat{Do}{do}{while}
\IncMargin{1em}
\begin{algorithm}[!t]
    \SetKwData{Left}{left}\SetKwData{This}{this}\SetKwData{Up}{up}
    \SetKwFunction{Union}{Union}\SetKwFunction{FindCompress}{FindCompress}
    \SetKwInOut{Input}{Input}\SetKwInOut{Output}{Output}\SetKwInOut{Notations}{Notations}
    \DontPrintSemicolon
    
    \Input{A classifier $C$, 
    a generator $G$, 
    the target class probabilities $y$,
    a convergence threshold $f_{min}$,
    the initial population size $t$,
    a latent space boundary $u$,
    the number of elite exemplars $k$ to keep during selection,
    the number of mutation operations $m$,
    a standard deviation $s$ for velocity sampling,
    a momentum rate $\alpha$.
    }
    \Notations{$z \in Z$ -- an exemplar $z$ from a population $Z$;
        $\mu \in M$ -- a momentum vector $\mu$ from a set $M$;
        $l$ -- the latent space size;\newline
        $v$ -- an $l$-dimensional velocity vector (perturbation applied to an exemplar during mutation).
    }
    \Output{Synthesized exemplar $z^*$.}
    \BlankLine

    \SetKwFunction{FMain}{mutate}
    \SetKwProg{Pn}{Function}{:}{\KwRet}
    \Pn{\FMain{$z$,$\mu$}}{
        $v \sim N(0, s)$\;
        \If{$\mu \neq \mathbf{0_{1,l}}$}{
            $v \gets \alpha \cdot \mu + (1 - \alpha) \cdot v$\;
        }
        $z' \gets z + v$\;
        \KwRet $(z', v)$
    }
    
    \SetKwFunction{FMain}{select}
    \SetKwProg{Pn}{Function}{:}{\KwRet}
    \Pn{\FMain{$Z,M$}}{
        $F \gets \emptyset$\;
        \For{$z_i \in Z$}{
            $f_i \gets \mathcal{L}(y, C(G(z_i))$\;
            $F \gets F \cup \{f_i\}$\;
        }        
        $I \gets argsort(F)$\;
        Z $\gets \{ z_{i_1}, z_{i_2}, ..., z_{i_k}\}$, where $i_j \in I$\;
        M $\gets \{ \mu_{i_1}, \mu_{i_2}, ..., \mu_{i_k}\}$, where $i_j \in I$\;
        \KwRet $(Z, M, \max_{f_i \in F} \{f_i\})$\;
    }
    \SetKwFunction{main}{main}
    \SetKwProg{myalg}{Algorithm}{}{}
        \myalg{\main{}}{
        $Z \gets \emptyset$\;
        $M \gets \emptyset$\;
        
        \For{$i \in \{1,2,...,t\}$}{
            $z_i \sim U(-u, u)$\;
            $Z \gets Z \cup \{z_i\}$\;
            $M \gets M \cup \{\mathbf{0_{1,l}}\}$\;
        }  
        
        $Z, M, f \gets select(Z,M)$\\
        \While{$f > f_{min}$}{
            \For{$i \in \{1,2,...,k\}$}{
                \For{$j \in \{1,2,...,m\}$}{
                    $z', \mu' \gets mutate(z_i,\mu_i)$\;
                    $Z \gets Z \cup \{z'\}$\;
                    $M \gets M \cup \{\mu'\}$\;
                }
            }
            $Z, M, f \gets select(Z,M)$\\
        }
        $z^* \gets z_1$\;
        \KwRet $z^*$\\
    }
\caption{Evolutionary Optimization Algorithm with Momentum}
\label{algorithm}
\end{algorithm}
\DecMargin{1em}

We hereby propose a novel evolutionary strategy based on optimization with momentum updates. We note that momentum is incorporated into a standard evolutionary strategy, i.e. the novelty consists in adding momentum updates. Our strategy is formally described in Algorithm \ref{algorithm}. In steps 17-22, our algorithm starts by sampling $t$ initial exemplars $z_i \in \mathbb{R}^l$ to form the initial population $Z$, such that $|Z|=t$. In step 20, each component of an exemplar $z_i$ is sampled from an uniform distribution $U$ over the interval $[-u,u]$. In the same time, we generate the set $M$ of momentum vectors $\mu_i$ associated to exemplars $z_i$. The initial momentum vectors are zero vectors of $l$ components, thus having the same size as the exemplars in $Z$. 

After initializing the population, we perform a selection in step 23 by keeping the top $k$ (elite) exemplars (and associated momentum vectors) that minimize our fitness function. The selection is performed inside the \emph{select} function defined in steps 7-15. The fitness of each exemplar in the current population $Z$ is computed in steps 8-11. Our fitness function is the sum squared error between the target output $y$ and the predicted output $\hat{y}$ for an exemplar $z_i \in Z$:
\begin{equation}\label{eq_loss}
\mathcal{L}(y, \hat{y}) = \sum_{j=1}^{n} (\hat{y}_j - y_j)^2,
\end{equation}
where $\hat{y}=C(G(z_i))$.

Until the fitness score $f$ of our least fit exemplar in $Z$ becomes smaller than $f_{min}$, we repeat steps 25-30. Inside the loop, each exemplar from the current population is duplicated and mutated $m$ times. The mutation, performed inside the \emph{mutate} function defined in steps 1-6, consists in adding a zero-centered Gaussian distributed velocity vector $v$ to the exemplar $z$. The mutation applied to the exemplar $z$ in step 5 can shift the new exemplar $z'$ in the direction of the gradient. We note that, during exemplar selection, we will choose exemplars that minimize our fitness function defined in Eq.~\eqref{eq_loss}. Since the kept exemplars were likely shifted in the right direction during the previous mutation, we added a momentum component to the mutation operation, which leads to faster convergence. The momentum vector $\mu$ is added to the velocity $v$ in a weighted sum computed in step 4, inside the \emph{mutate} function. The momentum $\mu$ is the previous perturbation (velocity) applied on the exemplar $z$. The magnitude of the momentum $\mu$ with respect to the generated velocity vector $v$ is controlled through the momentum rate $\alpha$.

The current population together with the mutated duplicates form a new population that passes through the selection process in step 30 of Algorithm \ref{algorithm}. By selecting only the top exemplars from every generation, we ensure that only the mutations that brought improvements are kept. Hence, only relevant perturbations are accumulated inside the momentum associated to an exemplar. 
As we will show in the experiments, the momentum component brings an increase of $19\%$ in convergence speed compared to the plain version (that does not use momentum), subject to using the same hyperparameters.
When the fitness of the least fit exemplar in $Z$ goes under the threshold $f_{min}$, we store the best exemplar $z_1$ in $z^*$ and return it as the output of our evolutionary algorithm.

\section{Experiments}
\label{sec_exp}
\subsection{Data Sets}


\noindent
{\bf Adult Data Set.}
For tabular data, we present experiments on the \textit{Adult Data Set} \cite{Kohavi-KDD-1996}. This is a binary classification data set for predicting the income of adults based on census information such as race, gender, marital status and level of education. It is composed of 48,842 samples with 14 features. 

\noindent
{\bf FER 2013.}
For image synthetization, we used the \textit{Facial Expression Recognition} (FER) 2013 \cite{Goodfellow-ICONIP-2013} data set that is comprised of grayscale images of faces representing 7 different classes of emotion. FER 2013 contains 28,709 training images, 3,589 validation images and 3,589 test images. The samples have a wide range of attributes, as they vary in illumination conditions, pose, gender, race and age. 

\noindent
{\bf Large Movie Review Dataset.}
We performed text synthetization experiments on the \textit{Large Movie Review Dataset} \cite{Maas-ACL-2011}. The training set contains 25,000 movie reviews for binary sentiment classification: positive or negative. The test set is similar in size. Each review is highly polarised, neutral or close-to-neutral samples being absent.

\subsection{Experimental Setup}

For Algorithm~\ref{algorithm}, we used the same hyperparameters in all experiments: the initial population size is $t=50$, the number of selected exemplars is $k=10$, the latent space boundary is $u=5$, the number of mutations per exemplar is $m=2$, the standard deviation used for mutations is $s=0.5$ and the momentum rate is $\alpha=0.3$. We present results with other hyperparameters in the supplementary.

\noindent
{\bf Setup for tabular data.}
To prove that our framework is truly model-agnostic, we employed a Random Forest (RF) classifier as the black box for the Adult Data Set, which attains an accuracy of $85.1\%$ on the test set, while being trained on half of the training set. In the pre-processing step, 4 of the 6 numerical features (except for capital-gain and capital-loss) were normalized and each of the 8 categorical features was passed through a different embedding layer, generating a vector of two components for each categorical feature. The concatenation of all these features ($6 + 8\cdot2 = 22$) gives us the final representation for the data samples. For data generation, we trained a VAE \cite{Eduardo-AISTATS-2020} on the other half of the training set (not used to train the RF classifier). We have synthesized prototype examples with regards to the output class probabilities of the RF classifier. The architecture of the VAE starts with two fully-connected layers having 64 and 128 neurons with batch normalization and Rectified Linear Unit (ReLU) activations, respectively. Finally, an 8-neuron dense layer determines the means and the standard deviations for a 4-dimensional encoding. During the reconstruction phase, embeddings are passed through two dense layers with 128 and 64 neurons, respectively. For the final output, there is an additional layer for numerical columns and an individual softmax layer for each categorical column. The loss is comprised of an $L_2$-distance component for numerical features and a categorical cross-entropy component for each categorical feature.

\noindent
{\bf Setup for image data.}
For facial expression recognition, we used the VGG-16 \cite{Simonyan-ICLR-2014b} architecture, which yields an accuracy of $67.4\%$ on the FER 2013 test set. As generators, we employed a Progressively Growing GAN \cite{Karras-ICLR-2018} and a VAE trained with cyclical annealing \cite{Fu-NAACL-2019}. The architectures and specifications for these networks are the ones specified in \cite{Karras-ICLR-2018} and \cite{Fu-NAACL-2019}, respectively. We performed several experiments on this data modality. Firstly, we provide a quantitative comparison between exemplars synthesized using our framework and exemplars generated in the glass-box scenario, i.e. when access to the classifier structure and weights is available, as in \cite{Nguyen-NIPS-2016}. Secondly, we provide a comparison of the convergence times of the two approaches, i.e. our evolutionary algorithm with momentum versus gradient descent (based on analytical gradients). In a set of preliminary trials, we noticed that the value of the gradient rapidly decreases within less than 5 iterations from values in the range of $10^{-1}$ to values in the range of $10^{-8}$, which impedes the gradient descent optimization process. Therefore, we experimented using gradient descent with momentum. We measure converge times from two perspectives: the number of model calls until the generated sample is classified with a confidence greater than $95\%$ and the duration of the optimization process in seconds. Thirdly, we show that our method is able to generate exemplars when the generator $G$ and the model $C$ are trained on the same data samples and on different data samples. Additionally, we prove that our method has the ability to generalize to previously unseen classes. To this end, we train the generator on all classes except one, e.g. \emph{surprised}, and successfully generate \emph{surprised} exemplars even though the generator has never seen a surprised face before.

\noindent
{\bf Setup for text data.} Sentence generation from latent embeddings has been proposed in the past, both through VAEs \cite{Bowman-CoNLL-2016,Chung-NIPS-2015} and through GANs \cite{Rajeswar-RepL4NLP-2017}. In our approach, we used an LSTM VAE \cite{Bowman-CoNLL-2016}, with a latent dimension of 128 neurons and hidden size of 512 neurons for both encoder and decoder networks. We used GloVe embeddings \cite{Pennington-EMNLP-2014} for the tokens processed by the encoder. The generator was trained for 120 epochs, with Kullback–Leibler annealing to avoid posterior collapse. The black-box classifier is a simple bidirectional LSTM, with hidden size of 256 neurons, with word embeddings trained alongside the final layer. The model achieves $85\%$ accuracy on the test set. The generator and the classifier are trained on disjoint training sets. Our method for manipulating text resembles those of \cite{Hu-ICML-2017,Yang-NIPS-2018}, since our classifier model acts as a sentiment discriminator. However, the classifier and generator networks are independent. Once they are established, we manipulate the generated text only by traversing the latent space.
\subsection{Results on Tabular Data}

\renewcommand{\arraystretch}{1.2}
\begin{table}[!t]
\centering
\caption{Exemplars synthesized by our evolutionary strategy with momentum for the Random Forest classifier trained on the Adult Data Set. There are two exemplars with \emph{low income} and two exemplars with \emph{high income}. Important features are highlighted in pale yellow.}
\label{tab:tabular}
\begin{tabular}{
>{\bfseries\arraybackslash}m{1.1in}
>{\arraybackslash}m{0.1in}
>{\arraybackslash}m{0.8in} >{\arraybackslash}m{0.8in}
>{\arraybackslash}m{0.12in}
>{\arraybackslash}m{0.8in} >{\arraybackslash}m{0.8in}
}
 \hline
 Features & & Low Income & Low Income & & High Income & High Income\\
 \hline
 \rowcolor{yellow!20}Age & &            28 &             19 & &           48&                 38\\
                     Work Class & &     Private &        Private & &      Private&            Private\\
                     Final weight & &         315124 &         393950 & &       105785&             45519\\
 \rowcolor{yellow!20}Education & &      5th-6th &        Some-college& &  Doctorate&          Prof-School\\
                     Educational-num& & 0 &              10 & &           19&                 15\\
                     Marital Status & & Never-married &  Never-married& & Married&            Married\\
                     Occupation & &     Other-service &  Other-service& & Prof-speciality&    Prof-speciality\\
                     Relationship & &   Other-relative & Own-child & &    Husband&            Husband\\
                     Race & &           White &          White & &        White&              White\\
 \rowcolor{yellow!20}Gender & &         Male &           Female & &       Male&               Male\\
                     Capital Gain & &   0 &              0 & &            0&                  0\\
                     Capital Loss & &   0 &              0 & &            0&                  0\\
 \rowcolor{yellow!20}Hours per Week& &  19 &             24 & &           64&                 84\\
 \rowcolor{yellow!20}Native Country & & Mexico &         United-States& & United-States&      United-States\\
 \hline
 \end{tabular}
 \end{table}
\renewcommand{\arraystretch}{1}

Synthesized tabular exemplars are not only meaningful by themselves, but they additionally provide a clear picture of the model's decision process. Even though the RF model is treated as a complete black-box, with no knowledge of its type or internal structure, we are able to deduce the model's reasoning by observing the synthesized prototypes in Table \ref{tab:tabular}. On the Adult Data Set, the features that influence the decision of the RF classifier seem to be the age, the level of education and the number of working hours per week. All \emph{high-income} exemplars are older people, with significant academic achievements (typically PhD) and more than 40 working hours per week. On the other end, \emph{low-income} exemplars have poor education, a young age and typically work part-time. Another type of \emph{low-income} exemplars (not included in Table \ref{tab:tabular}) features very old, retired and widowed people with poor education. Additionally, our analysis can reveal data set insights and model biases. In this scenario, the entirety of \emph{high-income} exemplars are males born in the United States, while \emph{low-income} exemplars are people born in Mexico. Hence, it seems that there is a bias towards classifying mexicans in the \emph{low income} category, which can raise ethical concerns towards racism.

\subsection{Results on Image Data}

In the image synthetization scenario, we present the differences and strengths of our framework when compared to a gradient descent approach that works in the glass-box scenario, while keeping the black-box scenario for our framework. For the comparison, we run both exemplar synthetization algorithms for 1000 times, generating 1000 exemplars in total. 

\begin{table}[!t]
\setlength\tabcolsep{3.2pt}
\centering
\caption{Convergence comparison between gradient descent with momentum, a standard evolutionary strategy and our evolutionary strategy with momentum on the FER 2013 data set. Reported values represent the number of runs in which the algorithms converged, the average number of model calls (forward passes) and the average time required to produce an exemplar. The times are measured in seconds on an NVidia GeForce RTX 2080 GPU with 8GB of RAM. Results are reported for 1000 runs.}
\label{tab:convergence_speed}
\begin{tabular}{lccc}
 \hline
 \textbf{Method}                                            & \textbf{Converged} & \textbf{Calls} & \textbf{Time}\\
                                                            & \textbf{(count)} & \textbf{(average)} & \textbf{(seconds)}\\

 \hline
 {Gradient descent with momentum} \cite{Nguyen-NIPS-2016}   & 955       & 378   & 3.77\\
 {Evolutionary strategy (ours)}                             & 1000      & 323   & 0.14\\
 {Evolutionary strategy with momentum (ours)}               & 1000      & 263   & 0.12\\
 \hline
 \end{tabular}

\end{table}
 
\noindent
{\bf Quantitative analysis.}
Considering the convergence results presented in Table~\ref{tab:convergence_speed}, we notice that both methods are able to synthesize exemplars that are classified with almost 100\% confidence. However, there is a significant difference between the two methods in terms of convergence. While our evolutionary strategy is able to converge each and every time, the gradient descent approach is highly dependant on its starting point. We observe that, for both GAN and VAE generators, the gradient descent optimization does not always converge. We found that, out of 1000 runs, in 45 of them the gradient descent with momentum method failed to synthesize an exemplar with over 95\% confidence, i.e. the algorithm got stuck in a non-optimal solution. This statement holds true when using both GANs and VAEs as generators.

\noindent
{\bf Running time.}
We measured the convergence times of the two exemplar generation approaches on an NVidia GeForce RTX 2080 GPU with 8GB of RAM. In Table \ref{tab:convergence_speed}, we present the number of model calls required to generate samples classified with more than $95\%$ confidence by the classifier. We also present the amount of physical time required for convergence. We observe that our evolutionary strategy with momentum requires fewer model calls (forward passes) than the gradient descent with momentum. In terms of physical time, our evolutionary strategy is about $31\times$ faster than gradient descent. We note that the gradient descent considered here is based on analytical gradients, which is faster than using numerical gradients. The remarkable difference in favor of our method can be explained by the following two factors: $(i)$ our evolutionary strategy is able to make model calls in batches and $(ii)$ it does not need to back-propagate gradients through the classifier or the generator. The experiments presented in Table \ref{tab:convergence_speed} also show the benefit of introducing momentum in the evolutionary strategy. In terms of model calls, the speed up brought by momentum is $19\%$.

\begin{figure}[!t]
    \centering
    \includegraphics[width=\linewidth]{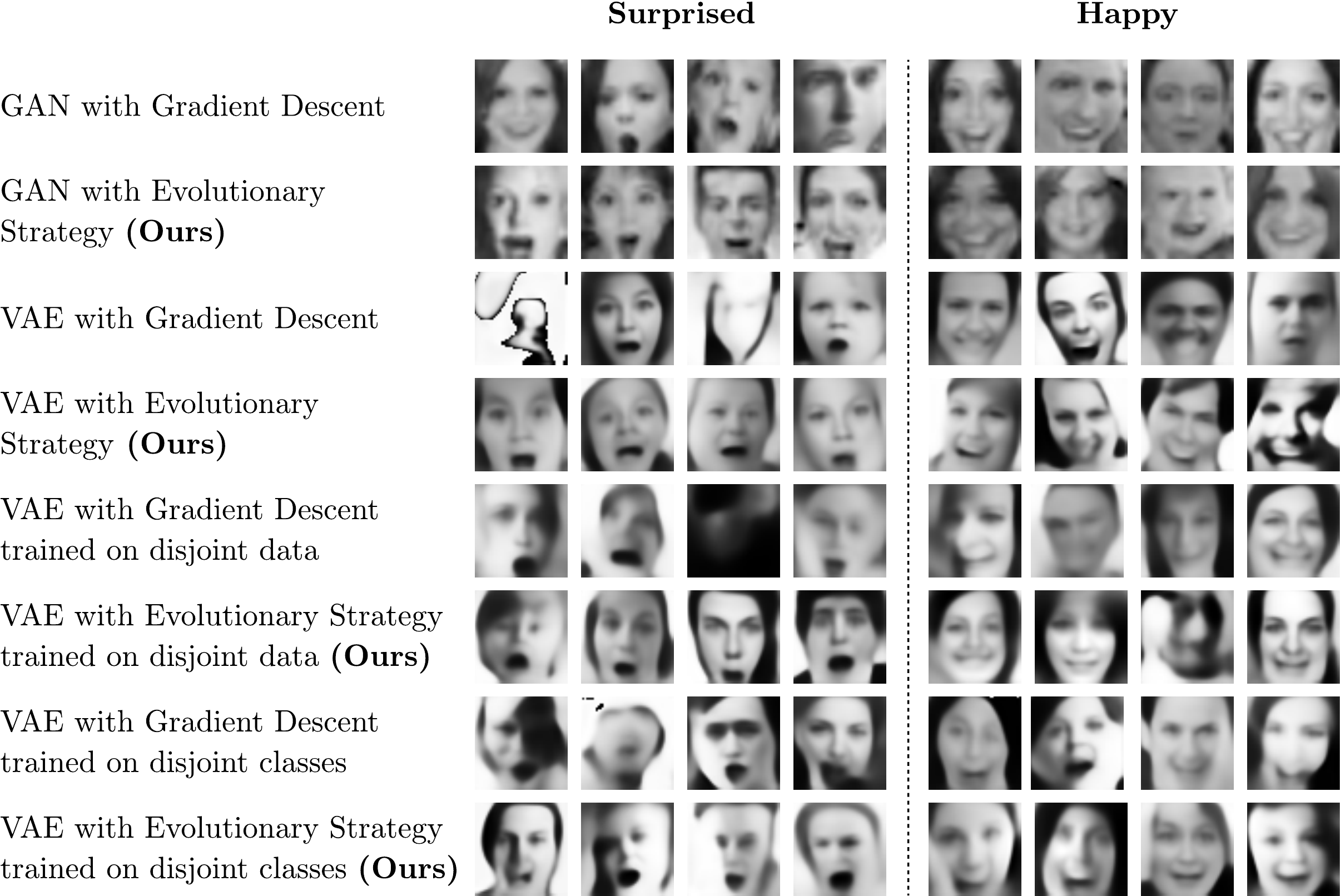}
    \caption{Exemplars synthesized for the VGG-16 neural network trained on the FER 2013 data set. Exemplars are generated in various scenarios (different generators, same training data, disjoint training data, disjoint classes) using our evolutionary strategy with momentum or gradient descent with momentum \cite{Nguyen-NIPS-2016}. In each scenario, we illustrate four exemplars for the \emph{surprised} class and another four exemplars for the \emph{happy} class.}
    \label{fig_image_exemplars}
\end{figure}

\noindent
{\bf Same training data.}
Since the GAN does not seem to produce realistic examples when its training data is not the same as that of the classifier, we present results in the context of using the same training data for both the generator and the classifier. We added this scenario to show that our method can produce exemplars with both GANs and VAEs. In Figure \ref{fig_image_exemplars}, we present a subset of representative exemplars for two classes: \emph{surprised} and \emph{happy}. When using a GAN as generator (first two rows in Figure \ref{fig_image_exemplars}), the exemplars present high quality visual features, irrespective of the synthetization algorithm. While the GAN exemplars are realistic, gradient descent does not always converge to a representative exemplar (the third \emph{happy} exemplar on first row in Figure \ref{fig_image_exemplars} seems \emph{neutral}). Moreover, gradient descent does not seem to always produce realistic exemplars for the VAE (see first and third exemplars for the \emph{surprised} class on third row in Figure~\ref{fig_image_exemplars}).

\noindent
{\bf Disjoint training data.}
We conducted experiments showing that our exemplar generation framework works well when the training data used for the generator is different from the training data used for the black-box classifier. The exemplars generated by our evolutionary strategy (sixth row in Figure \ref{fig_image_exemplars}) are still realistic and representative for the \emph{surprised} and \emph{happy} classes. The exemplars generated by gradient descent are not always realistic, and hard to interpret by humans (see third exemplar for the \emph{surprised} class on fifth row in Figure \ref{fig_image_exemplars}).

\noindent
{\bf Disjoint classes.}
We also conducted experiments to show that our exemplar synthetization framework generalizes to previously unseen classes. The images presented on the eighth row in Figure \ref{fig_image_exemplars} are generated with a VAE which was not trained on the respective classes,  \emph{surprised} and \emph{happy}. Still, the generated images seem realistic and representative for these two classes. Some exemplars produced by the gradient descent (seventh row in Figure \ref{fig_image_exemplars}) are less realistic.

\noindent
{\bf Summary.}
Considering the overall results, we notice that our evolutionary strategy does not get stuck in non-optimal solutions, while converging faster than gradient descent. Non-optimal solutions are avoided because the evolutionary strategy employs multiple starting points and the velocity values (used instead of gradients) always stay within a reasonable range. Since our method relies on making small jumps in the latent space, while ignoring the gradients, it can easily escape saddle points. The benefits of our evolutionary framework are empirically demonstrated by the results on FER 2013. In summary, we conclude that our method is more robust than gradient descent, while treating the classifier as a black-box. Indeed, we showed that access to the gradients or the training data distribution of the classifier is not required.

\subsection{Results on Text Data}

\renewcommand{\arraystretch}{1.2}
\begin{table}[!t]
\centering
\caption{Text exemplars synthesized by our evolutionary strategy with momentum for an LSTM classifier trained on the Large Movie Review Dataset. All samples are classified with high confidence (over $95\%$). Exemplars are provided for both polarity classes: \emph{positive} (left hand-side) and \emph{negative} (right hand-side).}
\label{tab:text}
\begin{tabular}{p{0.54\linewidth} | p{0.44\linewidth}}
 \hline
{\bf Positive Exemplars} & {\bf Negative Exemplars} \\
 \hline
\textit{``this is a great film and i recommend it to anyone."} & \textit{``one could have a cheap soap opera instead"} \\

\textit{``this is a great movie to watch and you will be a great time to watch it"} & \textit{``the film is not too long for the film to be a complete waste of time."} \\

\textit{``there is a lot of fun in this film and it is very well paced. "} & \textit{``the acting is not much to save the entire movie."} \\

\textit{``i am a fan of the genre but this is one of the best films of all time."} & \textit{``the final scene in the movie is the worst of the year."} \\

\textit{``this is a very good movie for everyone but it is not perfect."} & \textit{``the film is not a complete waste of time."} \\

\textit{``he a terrific actor and he is great as the lead and the performances are absolutely perfect."} & \textit{``it is not a terrible movie but it is not a bad film."}\\

\textit{``it is one of the best movies i have seen in a long time"} &  \textit{``the film is not that bad"} \\

\textit{``the film is well paced and it is not that good."} & \textit{``i mean it was not that bad"} \\

\textit{``it a good movie but it not worth a watch"} & \textit{``it was really worthless just below par."}\\
\hline
 \end{tabular}
 \end{table}
\renewcommand{\arraystretch}{1}

In Table \ref{tab:text}, we provide some selected exemplars generated for a simple LSTM text classifier, revealing the preferred inputs of the model for the \emph{positive} and \emph{negative} classes. We note that some generated reviews are realistic and representative for their class. Other reviews, especially the \emph{negative} ones, indicate that the classification model outputs wrong class probabilities with high confidence when it encounters some specific words. For example, sentences containing words such as ``good" or ``great" are classified as positive reviews, even though they appear in negated form, e.g. ``not that good". The classifier does not seem to understand contrasting transitions when evaluating the sentiment of reviews. These results are consistent with the problem of sentiment polarity classification observed by Li et al. \cite{Li-PACLIC-2009}. Hence, even though the classifier has a relatively high test accuracy ($85\%$), our method reveals that a naive training regime leads to sub-optimal results in real-world scenarios.

\section{Conclusion}
\label{sec_conclusion}

In this paper, we proposed a novel evolutionary strategy that incorporates momentum for generating exemplars for black-box models. Our framework requires an underlying generator, but it does not back-propagate gradients through the black-box model or the generator. We conducted experiments, showing that our approach can produce exemplars for three data types: image, text and tabular. Furthermore, our experiments indicate that our idea of incorporating momentum into a standard evolutionary strategy is useful, reducing the number of model calls by $19\%$. The empirical results demonstrate that our optimization algorithm converges faster than gradient descent with momentum, while providing similar or even more realistic exemplars. Given that our method does not require access to the weights or the training data of the black-box model, we believe it has a boarder applicability than gradient descent methods such as \cite{Nguyen-NIPS-2016}.

\bibliographystyle{splncs04}
\bibliography{refs.bib}

\begin{thebibliography}{10}
\providecommand{\url}[1]{\texttt{#1}}
\providecommand{\urlprefix}{URL }
\providecommand{\doi}[1]{https://doi.org/#1}

\bibitem{Adadi-Access-2018}
Adadi, A., Berrada, M.: {Peeking inside the black-box: A survey on Explainable
  Artificial Intelligence (XAI)}. IEEE Access  \textbf{6},  52138--52160 (2018)

\bibitem{Bien-AAS-2011}
Bien, J., Tibshirani, R.: Prototype selection for interpretable classification.
  The Annals of Applied Statistics pp. 2403--2424 (2011)

\bibitem{Bolukbasi-NIPS-2016}
Bolukbasi, T., Chang, K., Zou, J.Y., Saligrama, V., Kalai, A.: {Man is to
  Computer Programmer as Woman is to Homemaker? Debiasing Word Embeddings}. In:
  Proceedings of NIPS. pp. 4349--4357 (2016)

\bibitem{Bowman-CoNLL-2016}
Bowman, S.R., Vilnis, L., Vinyals, O., Dai, A.M., J{\'o}zefowicz, R., Bengio,
  S.: {Generating Sentences from a Continuous Space}. In: Proceedings of CoNLL.
  pp. 10–--21 (2016)

\bibitem{Caliskan-Science-2017}
Caliskan, A., Bryson, J.J., Narayanan, A.: Semantics derived automatically from
  language corpora contain human-like biases. Science  \textbf{356}(6334),
  183--186 (2017)

\bibitem{Chattopadhay-WACV-2018}
Chattopadhay, A., Sarkar, A., Howlader, P., Balasubramanian, V.N.: {Grad-CAM++:
  Generalized Gradient-Based Visual Explanations for Deep Convolutional
  Networks}. In: Proceedings of WACV. pp. 839--847 (2018)

\bibitem{Chen-NeurIPS-2019}
Chen, C., Li, O., Tao, D., Barnett, A., Rudin, C., Su, J.K.: {This Looks Like
  That: Deep Learning for Interpretable Image Recognition}. In: Proceedings of
  NeurIPS. pp. 8928--8939 (2019)

\bibitem{Chung-NIPS-2015}
Chung, J., Kastner, K., Dinh, L., Goel, K., Courville, A.C., Bengio, Y.: {A
  Recurrent Latent Variable Model for Sequential Data}. In: Proceedings of
  NIPS. pp. 2980--2988 (2015)

\bibitem{Cozma-ACL-2018}
Cozma, M., Butnaru, A., Ionescu, R.T.: Automated essay scoring with string
  kernels and word embeddings. In: Proceedings of ACL (2018)

\bibitem{Devlin-NAACL-2019}
Devlin, J., Chang, M.W., Lee, K., Toutanova, K.: {BERT: Pre-training of Deep
  Bidirectional Transformers for Language Understanding}. In: Proceedings of
  NAACL. pp. 4171--4186 (2019)

\bibitem{Eduardo-AISTATS-2020}
Eduardo, S., Nazábal, A., Williams, C.K.I., Sutton, C.: {Robust Variational
  Autoencoders for Outlier Detection and Repair of Mixed-Type Data}. In:
  Proceedings of AISTATS (2020)

\bibitem{Fu-NAACL-2019}
Fu, H., Li, C., Liu, X., Gao, J., Celikyilmaz, A., Carin, L.: {Cyclical
  Annealing Schedule: A Simple Approach to Mitigating KL Vanishing}. In:
  Proceedings of NAACL. pp. 240--250 (2019)

\bibitem{Georgescu-Access-2019}
Georgescu, M.I., Ionescu, R.T., Popescu, M.: Local learning with deep and
  handcrafted features for facial expression recognition. IEEE Access
  \textbf{7},  64827--64836 (2019)

\bibitem{Goodfellow-NIPS-2014}
Goodfellow, I., Pouget-Abadie, J., Mirza, M., Xu, B., Warde-Farley, D., Ozair,
  S., Courville, A., Bengio, Y.: Generative adversarial nets. In: Proceedings
  of NIPS. pp. 2672--2680 (2014)

\bibitem{Goodfellow-ICONIP-2013}
Goodfellow, I.J., Erhan, D., Carrier, P.L., Courville, A., Mirza, M., Hamner,
  B., Cukierski, W., Tang, Y., Thaler, D., Lee, D.H., Zhou, Y., Ramaiah, C.,
  Feng, F., Li, R., Wang, X., Athanasakis, D., Shawe-Taylor, J., Milakov, M.,
  Park, J., Ionescu, R.T., Popescu, M., Grozea, C., Bergstra, J., Xie, J.,
  Romaszko, L., Xu, B., Chuang, Z., Bengio, Y.: {Challenges in Representation
  Learning: A report on three machine learning contests}. In: Proceedings of
  ICONIP. vol.~8228, pp. 117--124 (2013)

\bibitem{Goodman-AIM-2017}
Goodman, B., Flaxman, S.: {European Union Regulations on Algorithmic Decision
  Making and a ``Right to Explanation''}. AI Magazine  \textbf{38}(3),  50--57
  (2017)

\bibitem{Guidotti-ECML-2019}
Guidotti, R., Monreale, A., Matwin, S., Pedreschi, D.: {Black Box Explanation
  by Learning Image Exemplars in the Latent Feature Space}. In: Proceedings of
  ECML--PKDD (2019)

\bibitem{Guidotti-CS-2018}
Guidotti, R., Monreale, A., Ruggieri, S., Turini, F., Giannotti, F., Pedreschi,
  D.: {A Survey Of Methods For Explaining Black Box Models}. ACM Computing
  Surveys  \textbf{51}(5),  1--42 (2018)

\bibitem{Gurumoorthy-ICDM-2019}
Gurumoorthy, K.S., Dhurandhar, A., Cecchi, G.A., Aggarwal, C.C.: In:
  Proceedings of ICMD. pp. 260--269 (2019)

\bibitem{He-CVPR-2016}
He, K., Zhang, X., Ren, S., Sun, J.: {Deep Residual Learning for Image
  Recognition}. In: Proceedings of CVPR. pp. 770--778 (2016)

\bibitem{Hu-ICML-2017}
Hu, Z., Yang, Z., Liang, X., Salakhutdinov, R., Xing, E.P.: {Toward Controlled
  Generation of Text}. In: Proceedings of ICML. pp. 1587--1596 (2017)

\bibitem{Karpathy-ICLR-2016}
Karpathy, A., Johnson, J., Li, F.: {Visualizing and Understanding Recurrent
  Networks}. In: Proceedings of ICLR (Workshop Track) (2016)

\bibitem{Karras-ICLR-2018}
Karras, T., Aila, T., Laine, S., Lehtinen, J.: {Progressive Growing of GANs for
  Improved Quality, Stability, and Variation}. In: Proceedings of ICLR (2018)

\bibitem{Kingma-ICLR-2014}
Kingma, D.P., Welling, M.: Auto-encoding variational bayes. In: Proceedings of
  ICLR (2014)

\bibitem{Kohavi-KDD-1996}
Kohavi, R.: {Scaling Up the Accuracy of Naive-Bayes Classifiers: a
  Decision-Tree Hybrid}. In: Proceedings of KDD. pp. 202--207 (1996)

\bibitem{Krizhevsky-NIPS-2012}
Krizhevsky, A., Sutskever, I., Hinton, G.E.: {ImageNet Classification with Deep
  Convolutional Neural Networks}. In: Proceedings of NIPS. pp. 1097--1105
  (2012)

\bibitem{LeCun-Nature-2015}
LeCun, Y., Bengio, Y., Hinton, G.: Deep learning. Nature  \textbf{521}(7553),
  436--444 (05 2015)

\bibitem{Li-NAACL-2016}
Li, J., Chen, X., Hovy, E., Jurafsky, D.: {Visualizing and Understanding Neural
  Models in NLP}. In: Proceedings of NAACL. pp. 681--691 (2016)

\bibitem{Li-PACLIC-2009}
Li, S., Huang, C.R.: {Sentiment Classification Considering Negation and
  Contrast Transition}. In: Proceedings of PACLIC. pp. 307--316 (2009)

\bibitem{Lundberg-NIPS-2017}
Lundberg, S.M., Lee, S.I.: {A Unified Approach to Interpreting Model
  Predictions}. In: Proceedings of NIPS. pp. 4765--4774 (2017)

\bibitem{Maas-ACL-2011}
Maas, A.L., Daly, R.E., Pham, P.T., Huang, D., Ng, A.Y., Potts, C.: {Learning
  Word Vectors for Sentiment Analysis}. In: Proceedings of ACL. pp. 142--150
  (2011)

\bibitem{Mahendran-CVPR-2015}
Mahendran, A., Vedaldi, A.: Understanding deep image representations by
  inverting them. In: Proceedings of CVPR. pp. 5188--5196 (2015)

\bibitem{Mahendran-IJCV-2016}
Mahendran, A., Vedaldi, A.: Visualizing deep convolutional neural networks
  using natural pre-images. International Journal of Computer Vision
  \textbf{120}(3),  233--255 (2016)

\bibitem{Mittelstadt-FAccT-2019}
Mittelstadt, B., Russell, C., Wachter, S.: {Explaining Explanations in AI}. In:
  Proceedings of FAccT. pp. 279--288. ACM (2019)

\bibitem{Mullenbach-NAACL-2018}
Mullenbach, J., Wiegreffe, S., Duke, J., Sun, J., Eisenstein, J.: {Explainable
  Prediction of Medical Codes from Clinical Text}. In: Proceedings of NAACL.
  pp. 1101--1111 (2018)

\bibitem{Nguyen-NIPS-2016}
Nguyen, A., Dosovitskiy, A., Yosinski, J., Brox, T., Clune, J.: Synthesizing
  the preferred inputs for neurons in neural networks via deep generator
  networks. In: Proceedings of NIPS. pp. 3387--3395 (2016)

\bibitem{Nguyen-CVPR-2015}
Nguyen, A.M., Yosinski, J., Clune, J.: {Deep Neural Networks are Easily Fooled:
  High Confidence Predictions for Unrecognizable Images} pp. 427--436 (2015)

\bibitem{Pennington-EMNLP-2014}
Pennington, J., Socher, R., Manning, C.D.: {GloVe: Global Vectors for Word
  Representation}. In: Proceedings of EMNLP. pp. 1532--1543 (2014)

\bibitem{Rajeswar-RepL4NLP-2017}
Rajeswar, S., Subramanian, S., Dutil, F., Pal, C.J., Courville, A.C.:
  {Adversarial Generation of Natural Language}. In: Proceedings of RepL4NLP.
  pp. 241–--251 (2017)

\bibitem{Ren-NIPS-2015}
Ren, S., He, K., Girshick, R., Sun, J.: {Faster R-CNN: Towards Real-Time Object
  Detection with Region Proposal Networks}. In: Proceedings of NIPS. pp. 91--99
  (2015)

\bibitem{Ribeiro-KDD-2016}
Ribeiro, M.T., Singh, S., Guestrin, C.: {"Why Should {I} Trust You?":
  Explaining the Predictions of Any Classifier}. In: Proceedings of KDD. pp.
  1135--1144 (2016)

\bibitem{Salimans-ArXiv-2017}
Salimans, T., Ho, J., Chen, X., Sutskever, I.: Evolution strategies as a
  scalable alternative to reinforcement learning. ArXiv
  \textbf{abs/1703.03864} (2017)

\bibitem{Selvaraju-ICCV-2017}
Selvaraju, R.R., Cogswell, M., Das, A., Vedantam, R., Parikh, D., Batra, D.:
  {Grad-CAM: Visual Explanations from Deep Networks via Gradient-based
  Localization}. In: Proceedings of ICCV. pp. 618--626 (2017)

\bibitem{Simonyan-ICLR-2014}
Simonyan, K., Vedaldi, A., Zisserman, A.: {Deep Inside Convolutional Networks:
  Visualising Image Classification Models and Saliency Maps}. In: Proceedings
  of ICLR (Workshop Track) (2014)

\bibitem{Simonyan-ICLR-2014b}
Simonyan, K., Zisserman, A.: {Very Deep Convolutional Networks for Large-Scale
  Image Recognition}. In: Proceedings of ICLR (2014)

\bibitem{Wiegreffe-EMNLP-2019}
Wiegreffe, S., Pinter, Y.: {Attention is not not Explanation}. In: Proceedingsa
  of EMNLP. pp. 11--20 (2019)

\bibitem{Yang-NIPS-2018}
Yang, Z., Hu, Z., Dyer, C., Xing, E.P., Berg{-}Kirkpatrick, T.: {Unsupervised
  Text Style Transfer using Language Models as Discriminators}. In: Proceedings
  of NIPS. pp. 7287--7208 (2018)

\bibitem{Yeh-NIPS-2018}
Yeh, C.K., Kim, J., Yen, I.E.H., Ravikumar, P.K.: Representer point selection
  for explaining deep neural networks. In: Proceedings of NIPS. pp. 9291--9301
  (2018)

\bibitem{Zhou-CVPR-2016}
Zhou, B., Khosla, A., Lapedriza, A., Oliva, A., Torralba, A.: {Learning Deep
  Features for Discriminative Localization}. In: Proceedings of CVPR. pp.
  2921--2929 (2016)

\end{thebibliography}

\newpage

\title{A Generic and Model-Agnostic Exemplar Synthetization Framework for Explainable AI - Supplementary Material} 
\author{}
\institute{}{}



\titlerunning{A Generic and Model-Agnostic Exemplar Synthetization Framework}
\tocauthor{A. Barbalau, A. Cosma, R.T. Ionescu, M. Popescu}
\toctitle{A Generic and Model-Agnostic Exemplar Synthetization Framework for Explainable AI}

\maketitle

\section*{Results with various hyperparameter settings}

We performed hyperparameter evaluation using the same image synthetization setup presented in the main article, showing how results can be influenced when parameters are optimized with regards to the given scenario. We have employed a VGG-16 \cite{Simonyan-ICLR-2014b} neural network classifier trained on the FER 2013 \cite{Goodfellow-ICONIP-2013} data set and Variational Autoencoder \cite{Fu-NAACL-2019} generator trained on the same data set. For each combination of parameters, we report the average number of model calls required to reach convergence over 1000 runs. Convergence is achieved when the exemplars that form the population are classified with over 95\% confidence by the black-box model. The corresponding results are presented in Table \ref{tab:hyper}. Here, $t$ represents the initial population, $k$ is the elite size, $\alpha$ represents the momentum, $m$ represents the number of mutations and $s$ the perturbation scale used for mutating each specimen. These results manage to showcase the individual properties of each parameter. Firstly, we can observe that as the initial population $t$ increases, the convergence time decreases, as expected. Secondly and more importantly, the elite size $k$ proved to have the most categorical impact on convergence speed. It can be seen that, as the elite size gets lower, the convergence speed drastically increases, as a result of the population focusing on mutating only the most promising specimens. Similar to the elite size, the number of mutations $m$ also presents a direct negative correlation with the convergence speed. Finally, the expected behavior from the momentum rate $\alpha$ and perturbation scale $s$ can also be seen in Table \ref{tab:hyper}. When there is little momentum or the perturbation is too small, the convergence time reaches its greatest values. As these parameters increase in value, the convergence speed increases as well and finally starts to decline when the values that become too high cause the population to overshoot.

\begin{table}[!t]
\setlength\tabcolsep{3.2pt}
\centering
\caption{Results with various hyperparameter settings for the Evolutionary Strategy with momentum applied on a VGG-16 classifier and a Variational Autoencoder trained on the FER 2013 data set. Here, $t$ represents the initial population, $k$ is the elite size, $\alpha$ represents the momentum, $m$ represents the number of mutations and $s$ the perturbation scale used for mutating each specimen. }
\label{tab:hyper}
\begin{tabular}{
>{\arraybackslash}m{0.6in}
>{\arraybackslash}m{.6in}
>{\arraybackslash}m{.6in} >{\arraybackslash}m{.6in}
>{\arraybackslash}m{.6in} >{\centering\arraybackslash}m{1in}
}
 \hline
 $\boldsymbol{t}$	& $\boldsymbol{k}$ & $\boldsymbol{\alpha}$ &$\boldsymbol{m}$ & $\boldsymbol{s}$ & \textbf{\#Model calls}\\

 \hline
 
 50 & 10 & 0.3 & 2 & 0.5 & 263.2\\
 \hline
 
 30 & 10 & 0.3 & 2 & 0.5 & 277.6\\
 40 & 10 & 0.3 & 2 & 0.5 & 247.5\\
 60 & 10 & 0.3 & 2 & 0.5 & 251.2\\
 70 & 10 & 0.3 & 2 & 0.5 & 229.6\\
 \hline
 
 50 & 5 & 0.3 & 2 & 0.5 & 167.4\\
 50 & 15 & 0.3 & 2 & 0.5 & 336.5\\
 50 & 20 & 0.3 & 2 & 0.5 & 458.4\\
 50 & 25 & 0.3 & 2 & 0.5 & 537.3\\
 50 & 30 & 0.3 & 2 & 0.5 & 639.6\\
 \hline
 
 50 & 10 & 0   & 2 & 0.5 & 323.0\\
 50 & 10 & 0.1 & 2 & 0.5 & 275.3\\
 50 & 10 & 0.2 & 2 & 0.5 & 274.8\\
 50 & 10 & 0.4 & 2 & 0.5 & 260.1\\
 50 & 10 & 0.5 & 2 & 0.5 & 256.3\\
 50 & 10 & 0.6 & 2 & 0.5 & 251.2\\
 50 & 10 & 0.7 & 2 & 0.5 & 237.3\\
 50 & 10 & 0.8 & 2 & 0.5 & 299.3\\
 50 & 10 & 0.9 & 2 & 0.5 & 408.0\\
 \hline
 
 50 & 10 & 0.3 & 1 & 0.5 & 182.8\\
 50 & 10 & 0.3 & 3 & 0.5 & 293.3\\
 50 & 10 & 0.3 & 4 & 0.5 & 358.7\\
 50 & 10 & 0.3 & 5 & 0.5 & 429.4\\
 \hline
 
 50 & 10 & 0.3 & 2 & 0.1 & 318.9\\
 50 & 10 & 0.3 & 2 & 0.2 & 286.7\\
 50 & 10 & 0.3 & 2 & 0.3 & 274.8\\
 50 & 10 & 0.3 & 2 & 0.4 & 233.5\\
 50 & 10 & 0.3 & 2 & 0.6 & 263.6\\
 50 & 10 & 0.3 & 2 & 0.7 & 278.2\\
 50 & 10 & 0.3 & 2 & 0.8 & 303.4\\
 50 & 10 & 0.3 & 2 & 0.9 & 297.6\\
 \hline

 \end{tabular}
\end{table}

\section*{Exemplar synthetization on high resolution images}

\begin{figure}[!t]
    \centering
    \includegraphics[width=\linewidth]{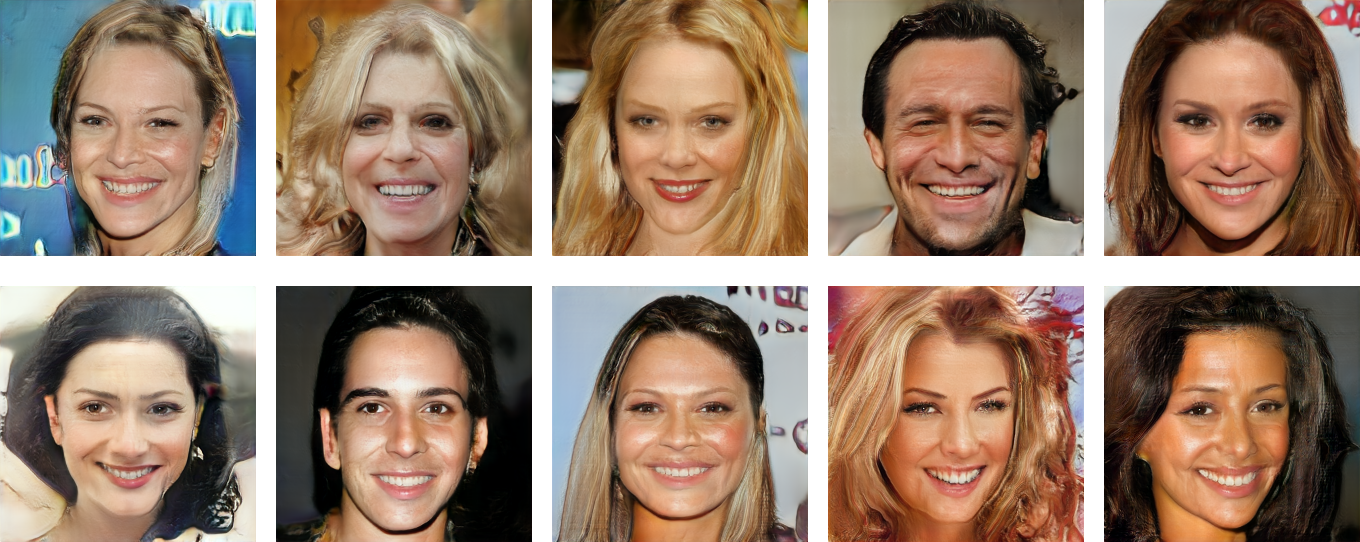}
    \caption{Exemplars synthesized for the \emph{happy} class for a VGG-16 neural network trained on the FER 2013 data set using a pre-trained CelebA-HQ Progressively Growing GAN that produces high quality 512$\times$512 images.}
    \label{fig_smile}
\end{figure}

\begin{figure}[!t]
    \centering
    \includegraphics[width=\linewidth]{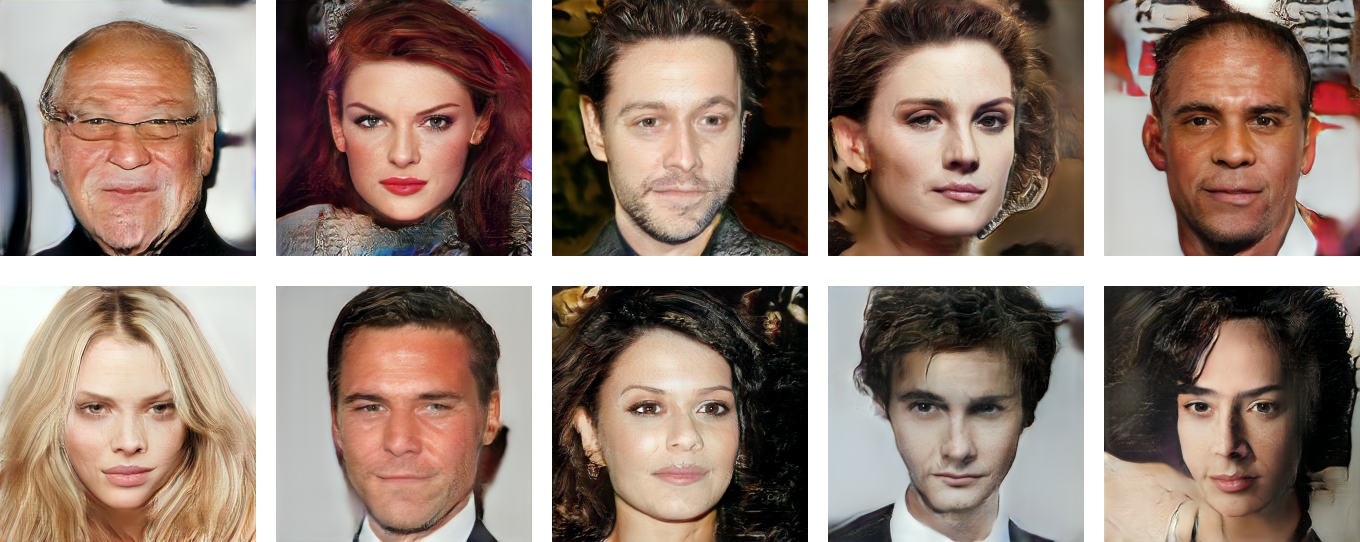}
    \caption{Exemplars synthesized for the \emph{neutral} class for a VGG-16 neural network trained on the FER 2013 data set using a pre-trained CelebA-HQ Progressively Growing GAN that produces high quality 512$\times$512 images.}
    \label{fig_neutral}
\end{figure}

In order to test our framework further, we have employed a pre-trained CelebA-HQ Progressively Growing GAN \cite{Karras-ICLR-2018} as a generator for the facial expression recognition black-box model used in the previous experiments. This generator produces high quality 512$\times$512 images that come from a completely different distribution than that of FER 2013 samples. Considering this, when performing the optimization, the generated samples were preprocessed to fit the size and the color requirements when passed to the classifier. Even though the prediction model and the generator were trained on two different data sets, the framework was able to synthesize realistic and high quality exemplars. Figures \ref{fig_smile} and \ref{fig_neutral} show results for the \emph{happy} and \emph{neutral} classes, respectively. This experiment shows that our framework can also synthesize high-resolution color images, even if the black-box classifier takes as input low-resolution grayscale images.

\end{document}